\documentclass[preprint,12pt,num]{elsarticle}

\usepackage{geometry}
\usepackage{amsmath}
\usepackage{amssymb}
\usepackage{amsthm}
\usepackage{booktabs}

\usepackage[T1]{fontenc}
\usepackage[utf8]{inputenc}
\usepackage{color}
\usepackage{array}
\usepackage{verbatim}
\usepackage{rotfloat}
\usepackage{url}
\usepackage{tabularx} 
\usepackage{algorithmicx}
\usepackage{algpseudocode}
\usepackage{algorithm}
\usepackage{setspace}
\usepackage{enumitem} 
\usepackage{subcaption} 
\usepackage{wrapfig}
\usepackage{threeparttable}
\usepackage{rotating}
\usepackage{pstricks, pst-text, pst-node, pst-plot}
\usepackage{epsfig}
\usepackage[english]{babel}
\usepackage{caption}

\definecolor{note_fontcolor}{rgb}{0.800781, 0.800781, 0.800781}

\journal{Chemometrics and Intelligent Laboratory Systems}

\begin{document}

\begin{frontmatter}

\title{Feature Space Selection and Heterogeneous Effect Estimation for Blood-Brain Barrier Permeability: A Random Forest to the Generalized Random Forest Pipeline}

\author[1]{Tshemollo Rapolai\corref{cor1}}
\ead{u25747445@tuks.co.za}

\author[1]{S. L. Makgai}
\ead{seite.makgai@up.ac.za}

\author[2,1]{M. Arashi}
\ead{mohammad.arashi@up.ac.za}
\ead{arashi@um.ac.ir}

\cortext[cor1]{Corresponding author}

\affiliation[1]{organization={Department of Statistics, University of Pretoria},
            city={Pretoria},
            country={South Africa}}

\affiliation[2]{organization={Department of Statistics, Faculty of Mathematical Sciences, Ferdowsi University of Mashhad},
            addressline={P.O. Box 1159},
            city={Mashhad},
            postcode={91775},
            country={Iran}}

\begin{abstract}
Predicting blood-brain barrier (BBB) permeability is critical for central nervous system drug discovery. Using the MoleculeNet BBBP dataset ($n=2039$), this study systematically ablates molecular feature spaces to isolate featurisation from model architecture. We evaluate three feature families (Morgan fingerprints, RDKit physicochemical descriptors, SMILES bigrams) across four learning algorithms. Results demonstrate that predictive performance depends jointly on feature representation and algorithm. Dynamic Random Forest using combined features achieved the highest mean AUC (0.970, 95\% CI: 0.963-0.977). 

Second, this optimal representation enables exploratory estimation of heterogeneous associations between molecular structure and BBB permeability using Generalized Random Forests. Constructing a pseudo-treatment from a LogP median split, we applied double/debiased machine learning to account for confounding. Orthogonalization substantially attenuates the heterogeneity detected by naive causal forests; no conditional effects remained significant after false discovery rate correction (smallest adjusted $p=0.082$). Furthermore, orthogonalized feature importance shifted toward residual structural information in SMILES bigrams.

Ultimately, once observed confounding is properly accounted for, evidence that LogP-BBB associations vary systematically across chemical space is insufficient. This underscores that feature representation and model architecture are coupled design choices, and that unorthogonalized causal forests risk overstating genuine treatment-effect heterogeneity.
\end{abstract}

\end{frontmatter}

\section{Introduction}\label{sec:introduction}

The blood-brain barrier (BBB) poses a significant challenge in central nervous system drug discovery, as it restricts the passage of many therapeutic compounds from the systemic circulation into the brain parenchyma \citep{Pardridge2005}. Accurate prediction of BBB permeability is therefore essential for prioritizing lead compounds and reducing late-stage attrition. The BBBP dataset has emerged as a widely adopted benchmark for developing computational models to distinguish permeable from impermeable compounds \citep{Martins2012}.

The prediction task is formulated as a binary classification problem in which each molecule is assigned a label based on its experimentally determined permeability. Due to the class imbalance typically observed in such data, AUC-ROC was selected as the primary evaluation metric, since it provides a threshold-independent assessment of classifier performance and is less sensitive to imbalance than overall accuracy.

A range of classical machine learning algorithms have been applied to BBB permeability prediction, most commonly Random Forests and Support Vector Machines, owing to their robustness to moderate sample sizes and their ability to model non-linear relationships between molecular descriptors and permeability \citep{Martins2012, Muehlbauer2023}. Ensemble tree-based methods, in particular, have consistently outperformed linear classifiers on this task, though the relative contribution of the underlying feature representation to this advantage is rarely disentangled from the choice of algorithm itself.

Most computational studies of BBB permeability stop at this predictive question, treating the compound as a fixed point for classification rather than asking whether its molecular properties bear a uniform relationship to permeability across chemical space. This distinction matters: a property such as lipophilicity may govern permeability differently depending on the scaffold, polarity, or hydrogen-bonding profile of the surrounding molecule, and an estimate of its average association across the full dataset can mask this variation entirely. Estimating such heterogeneity from observational data of this kind is not straightforward, however, since molecular descriptors that influence permeability are frequently correlated with the property under investigation. A naive analysis risks attributing to heterogeneity what is in fact residual confounding.

A number of frameworks have been developed in the causal machine learning literature to address this challenge. The X-learner \citep{Kunzel2019} estimates heterogeneous effects by combining separate outcome models across treatment arms, while the DR-learner \citep{Kennedy2023} constructs a doubly robust pseudo-outcome that remains consistent if either the outcome or treatment model is correctly specified. OrthoForest \citep{pmlr-v97-oprescu19a} extends the generalized random forest framework of \cite{Athey2019} by incorporating Neyman-orthogonal moment conditions directly into the forest-growing procedure, reducing sensitivity to nuisance parameter estimation error. Building on this principle, the present study employs an orthogonalized causal forest that residualizes both the outcome and the treatment on the covariate space prior to effect estimation, thereby limiting the extent to which observed confounding is misattributed to genuine heterogeneity.

This study therefore pursues two aims: (i) identify a molecular feature representation that balances predictive performance with computational efficiency, and (ii) explore heterogeneous associations between molecular lipophilicity and BBB permeability using a Generalized Random Forest, a flexible framework for estimating localized statistical parameters from observational data. The two aims are sequential rather than independent: the representation identified in (i) provides the covariate space in which the heterogeneity analysis in (ii) is conducted. Accordingly, this paper first conducts a systematic ablation of molecular feature space (Section \ref{sec2}) to identify a parsimonious representation, before applying an orthogonalized Generalized Random Forest to the resulting feature space to estimate heterogeneous associations with lipophilicity (Sections \ref{sec3}-\ref{sec6}).

\section{Ablation Study}\label{sec2}
This section addresses aim (i): identifying a molecular feature representation that balances predictive performance with computational efficiency.

\subsection{Task and Dataset}
We address a binary classification problem: predicting whether a small molecule can penetrate the blood-brain barrier (BBB) using the BBBP dataset from the MoleculeNet benchmark \citep{Wu2018}. After removing invalid SMILES entries, the dataset contains 2,039 molecules, each encoded as a SMILES string \citep{Weininger1988} and labeled as BBB-permeable ($y = 1$) or non-permeable ($y = 0$). The dataset is imbalanced (76\% positive, 24\% negative), reflecting the composition of the benchmark rather than an assumed population distribution; we therefore report AUC alongside F1 score and balanced accuracy rather than relying on raw accuracy alone. SMILES strings serve as the sole input to our feature-engineering pipeline, with no 3D coordinates, molecular graphs, or precomputed descriptors used at the input stage.

\subsection{Molecular Feature Spaces}
The central goal of this study is to isolate the contribution of different molecular feature families to BBB permeability prediction. We construct three feature families directly from SMILES strings using standard cheminformatics and NLP libraries, with no hand-crafted rules or dataset-specific tuning, ensuring the pipeline is reproducible on any SMILES-encoded dataset.

\begin{itemize}
    \item \textbf{Morgan fingerprints (M).} A 2048-bit binary ECFP4 fingerprint (radius 2) \citep{Rogers2010}, computed via RDKit \citep{Landrum2023}, encoding local chemical environments up to two bonds from each heavy atom. This representation captures local connectivity and the presence of substructures but discards global molecular properties and count information.
    \item \textbf{RDKit descriptors (D).} Twenty continuous physicochemical descriptors spanning molecular size (molecular weight, heavy atom count), lipophilicity ($\log P$), polarity (TPSA, hydrogen-bond donor/acceptor counts), flexibility (rotatable bonds, ring statistics), and electronic properties (partial charges). These descriptors provide a global physicochemical context absent from fingerprint-based representations and are known to influence BBB penetration \citep{Wager2010}.
    \item \textbf{SMILES bigrams (B).} TF-IDF-weighted character bigrams extracted directly from raw SMILES strings, following NLP-based molecular representation approaches \citep{Jardim2024}. Unlike fingerprint- or descriptor-based features, this representation preserves the raw syntax of SMILES notation (branching, bond symbols, ring closures) without imposing a predefined chemical vocabulary.
\end{itemize}

These three families are combined into seven feature spaces spanning all singleton, pairwise, and full combinations of M, D, and B (Table \ref{tab1}), enabling systematic evaluation of the standalone sufficiency, pairwise synergy, and redundancy of each family.

\subsection{Classifiers}
We evaluate four classifiers spanning linear and non-linear, fixed-complexity and adaptive-complexity model classes: Logistic Regression ($\ell_2$-regularized), Random Forest \citep{Breiman2001} with 500 trees, and a Support Vector Machine with an RBF kernel \citep{Cortes1995}, all implemented via scikit-learn \citep{scikit-learn} with default hyperparameters unless otherwise stated.

In addition, we include the Dynamic Random Forest (DRF) \citep{Bernard2012}, which removes the requirement to pre-specify the number of trees $T$. Trees are added incrementally, and out-of-bag (OOB) error is monitored after each addition; training terminates automatically once OOB performance fails to improve by more than a tolerance $\delta$ for $p$ consecutive rounds. This data-adaptive stopping criterion avoids both underfitting and unnecessary computation, which is particularly valuable in the small-sample regime characteristic of early-phase drug discovery.

\begin{figure}
    \centering
    \includegraphics[width=0.75\linewidth]{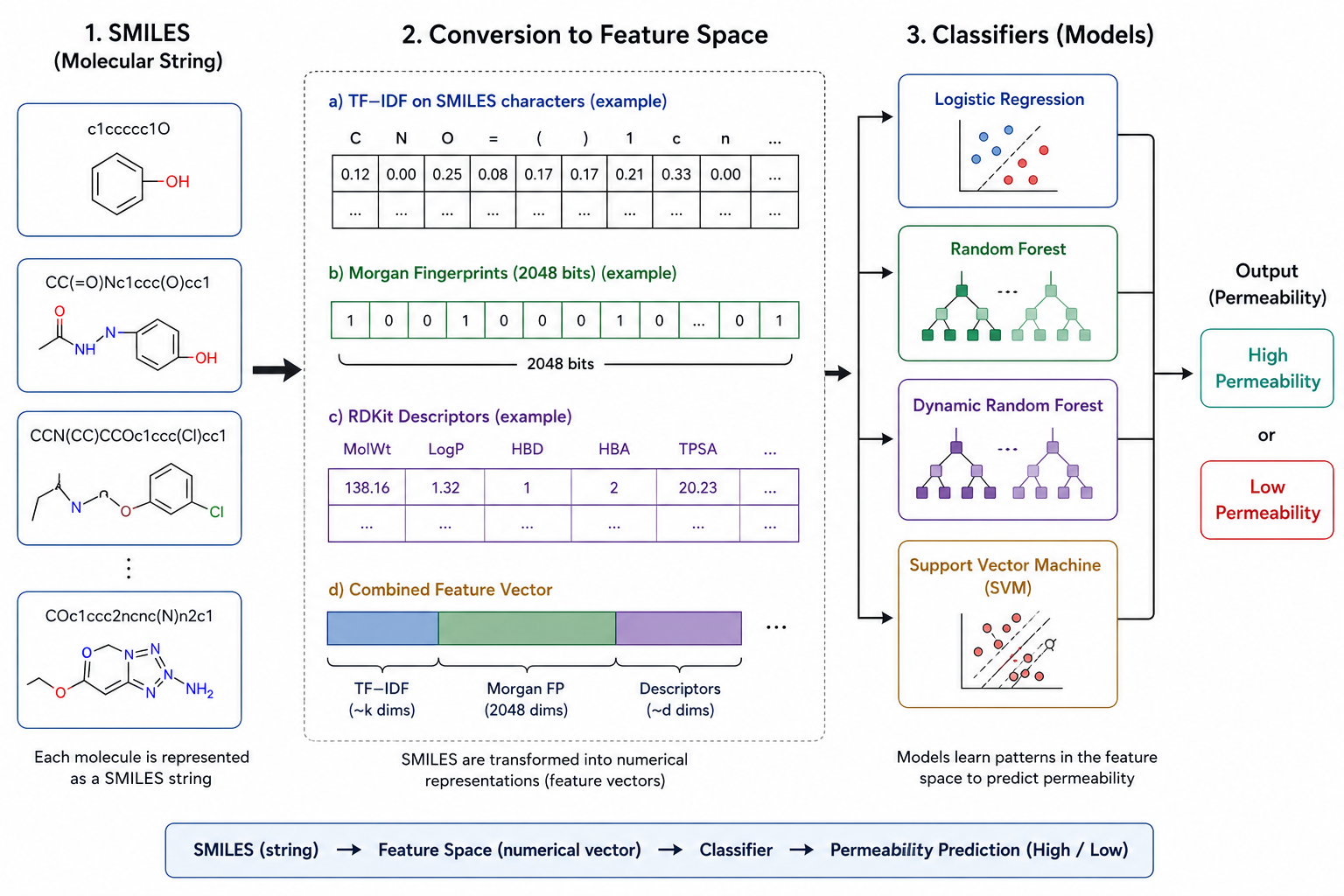}
    \caption{Overview of the SMILES-based permeability prediction pipeline.}
    \label{fig:placeholder}
\end{figure}

\subsection{Bootstrap Evaluation Protocol}
Feature representation, sample size, and learning algorithm are varied independently under a stratified bootstrap simulation protocol, decoupling three sources of variation: feature representation (M, D, B, and their combinations), sample size (1631), and classifier (LR, RF, DRF, SVM). For each sample size, $B = 1000$ stratified bootstrap samples are drawn from the training set, each classifier is trained and evaluated on a fixed held-out test set, and mean AUC, variance, 95\% confidence intervals, and relative efficiency are computed across replications (Algorithm \ref{algo1}).

\begin{algorithm}
\caption{Stratified Bootstrap Simulation Study with Multiple Classifiers}\label{algo1}
\begin{algorithmic}[1]
\Require Training data $(X_{\text{train}}, y_{\text{train}})$, fixed test set $(X_{\text{test}}, y_{\text{test}})$, sample sizes $N$, replications $B$, classifiers $C = \{\text{LR, RF, DRF, SVM}\}$
\Ensure Mean AUC, variance, confidence intervals, relative efficiency, statistical tests
\State Compute class proportions $\pi_n \in N$
\State Compute stratified sizes $n_+$ and $n_-$
\For{$b = 1$ to $B$}
    \State Draw stratified bootstrap sample $S^{(b)}$
    \For{$c \in C$}
        \State Fit model $c$ on $S^{(b)}$
        \State Predict probabilities on $X_{\text{test}}$
        \State Compute AUC: $\text{AUC}^{(b)}_c$
    \EndFor
\EndFor
\State Compute pairwise performance differences (e.g., DRF vs. LR)
\State Aggregate mean AUC, variance, and 95\% confidence intervals over $B$ replications
\State Compute relative efficiency measures
\end{algorithmic}
\end{algorithm}

\subsection{Results}
Table \ref{tab1} reports mean AUC (95\% bootstrap CI) across all seven feature spaces and four classifiers at the full training sample size ($n = 1631$, $B = 1000$ replications).

\begin{table}[h]
\centering
\caption{Mean AUC (95\% bootstrap CI) across seven feature spaces and four classifiers ($n = 1631$, $B = 1000$).}\label{tab1}
\resizebox{\textwidth}{!}{%
\begin{tabular}{l c c c c}
\toprule
Feature Space & LR & RF & DRF & SVM \\
\midrule
B & 0.948 (0.942-0.955) & 0.960 (0.951-0.967) & 0.959 (0.950-0.966) & 0.940 (0.926-0.952) \\
D & 0.887 (0.875-0.898) & 0.927 (0.913-0.941) & 0.925 (0.911-0.939) & 0.910 (0.897-0.923) \\
M & 0.906 (0.886-0.923) & 0.927 (0.913-0.940) & 0.926 (0.910-0.940) & 0.900 (0.882-0.918) \\
M+D & 0.913 (0.896-0.928) & 0.947 (0.936-0.956) & 0.946 (0.933-0.956) & 0.903 (0.886-0.920) \\
M+B & 0.932 (0.917-0.946) & 0.968 (0.961-0.974) & 0.967 (0.959-0.973) & 0.924 (0.910-0.939) \\
D+B & 0.950 (0.942-0.956) & 0.968 (0.962-0.974) & 0.967 (0.961-0.974) & 0.945 (0.932-0.956) \\
M+D+B & 0.936 (0.924-0.948) & 0.972 (0.966-0.977) & 0.970 (0.963-0.977) & 0.926 (0.912-0.941) \\
\bottomrule
\end{tabular}%
}
\end{table}

Several patterns emerge from the ablation. Among the three individual feature families, SMILES bigrams (B) provide the strongest standalone signal (RF AUC 0.960), outperforming both Morgan fingerprints (M, 0.927) and RDKit descriptors (D, 0.927) despite requiring no explicit chemical domain knowledge. Descriptors alone are the weakest individual representation, consistent with their role as summary physicochemical statistics rather than structural encodings.

Pairwise combination reveals that bigrams, rather than any specific pairing of feature families, drive most of the gain over single-family representations. Both pairwise combinations that include bigrams---D+B (RF AUC 0.968) and M+B (RF AUC 0.968)---achieve performance statistically indistinguishable from one another and approach the full three-family combination, whereas M+D, the only pairwise combination that excludes bigrams, yields a substantially smaller gain (RF AUC 0.947). This asymmetry indicates that SMILES bigram features carry information largely complementary to that of both Morgan fingerprints and physicochemical descriptors, whereas fingerprints and descriptors alone are comparatively redundant with one another for this task. The full combined space (M+D+B) achieves the highest point estimate overall (RF AUC 0.972), but the improvement over either bigram-containing pairwise space is small relative to the width of the confidence intervals.

Across representations, tree-based ensembles (RF, DRF) consistently outperform Logistic Regression and SVM, and RF and DRF track each other closely throughout, differing by no more than 0.002 AUC in any condition. Logistic Regression is the only classifier to degrade under feature augmentation, achieving its best performance under the bigram-only representation (0.948) rather than the full feature space (0.936), consistent with multicollinearity and non-linear interactions among the combined features that are incompatible with a linear decision boundary. SVM underperforms tree-based methods across all seven feature spaces, consistent with its known sensitivity to high-dimensional correlated features.

\subsection{Feature Space Selection for HTE Analysis}
Although M+D+B achieved the highest point estimate of the mean AUC under Random Forest, a direct paired comparison against D+B, computed per replication across the same $B = 1000$ bootstrap draws, found no statistically significant difference between the two spaces. The mean paired difference in AUC (M+D+B $-$ D+B) was 0.003 (95\% CI: $-0.002-0.009$, $p = 0.204$) under Random Forest and 0.003 (95\% CI: $-0.004-0.010$, $p = 0.386$) under Dynamic Random Forest; in both cases the confidence interval for the difference includes zero.

Given this statistical equivalence, the D+B feature space combined with Random Forest was selected for the subsequent heterogeneity analysis (Sections \ref{sec3}-\ref{sec6}). This choice is preferred on grounds of parsimony: D+B reduces the feature dimensionality from over 2000 (when Morgan fingerprints are included) to 269 dimensions (20 physicochemical descriptors and 249 SMILES bigram TF-IDF features), an approximately eightfold reduction. This lower-dimensional representation improves the interpretability of feature-importance measures within the causal forest and reduces the covariate dimensionality entering the orthogonalization step, without incurring a measurable cost to predictive performance.

\subsection{Robustness to Scaffold-Based Splitting}
The results in Table \ref{tab1} are obtained under a random 80/20 split of the 2,039-molecule dataset. Random splitting can overstate generalization performance in molecular property prediction, since structurally similar molecules---sharing a common scaffold---may be distributed across both training and test sets, allowing models to exploit local chemical similarity rather than learning transferable structure-activity relationships \citep{bemis1996properties}. To assess whether the feature space comparisons above hold under a more demanding generalization test, we repeat the evaluation using Bemis-Murcko scaffold splitting \citep{bemis1996properties}, in which molecules are grouped by scaffold and whole scaffold groups---rather than individual molecules---are assigned to training or test sets, following the greedy protocol used in MoleculeNet \citep{Wu2018}.

Scaffold decomposition of the 2,039-molecule dataset yields 1,102 distinct scaffold groups, of which 843 (76.5\%) are singletons (a unique scaffold represented by exactly one molecule) and the largest group contains 137 molecules. To characterize variability under this stricter split, we generate 200 independent scaffold splits by reshuffling scaffold-group order prior to the greedy 80/20 fill, refitting all feature extractors and classifiers within each split to prevent information leakage from the test scaffolds into training-derived transformations (e.g. the TF-IDF vocabulary for bigram features).

Table \ref{tab2} reports mean AUC and 95\% percentile CIs across the 200 scaffold splits, restricted to Random Forest and Dynamic RF, the two best-performing classifiers under random splitting.

\begin{table}[h]
\caption{Mean AUC (95\% CI) across 200 scaffold splits, Random Forest and Dynamic RF only.}\label{tab2}
\begin{tabular*}{\textwidth}{@{\extracolsep\fill}lll}
\toprule
Feature Space & RF & DRF \\
\midrule
B & 0.9465 (0.9328-0.9602) & 0.9458 (0.9325-0.9605) \\
D & 0.9002 (0.8757-0.9231) & 0.8989 (0.8713-0.9213) \\
M & 0.9012 (0.8741-0.9289) & 0.8994 (0.8699-0.9285) \\
M+D & 0.9247 (0.9020-0.9455) & 0.9236 (0.9009-0.9452) \\
M+B & 0.9542 (0.9400-0.9671) & 0.9534 (0.9400-0.9666) \\
D+B & 0.9551 (0.9428-0.9681) & 0.9541 (0.9411-0.9674) \\
M+D+B & 0.9591 (0.9470-0.9735) & 0.9582 (0.9455-0.9717) \\
\bottomrule
\end{tabular*}
\end{table}

Under scaffold splitting, mean AUC for the best-performing configuration (M+D+B, RF) falls from 0.972 to 0.959---a decline of approximately one percentage point, modest relative to degradations of 5-10 points typically reported for scaffold-split evaluation of learned molecular representations \citep{Wu2018}. Critically, the central finding of the random-split ablation is preserved under this harder test: both bigram-containing pairwise spaces, D+B (0.9551, CI 0.9428-0.9681) and M+B (0.9542, CI 0.9400-0.9671), remain statistically indistinguishable from the full M+D+B combination (0.9591, CI 0.9470-0.9735), while M+D---the pairwise space excluding bigrams---again lags meaningfully behind (0.9247, CI 0.9020-0.9455). This confirms that the role of bigram features as the primary driver of near-ceiling performance is not an artifact of the random splitting protocol, but persists under scaffold-based generalization.

The relative ordering of feature spaces is fully preserved between splitting protocols: single-family representations (M, D alone) remain weakest, bigrams (B) remain the strongest standalone family, and pairwise/full combinations retain the same rank ordering observed under random splitting. Tree-based ensembles again track each other closely, differing by no more than 0.001-0.002 AUC in any condition, consistent with the random-split results. Confidence interval widths are correspondingly wider under scaffold splitting (0.0265 for M+D+B/RF versus 0.011 under random splitting, an approximately 2.4-fold increase), reflecting the smaller effective number of independent scaffold groups (1,102) relative to the number of individual molecules available for random resampling---a property of the harder, more honest evaluation protocol rather than a weakness of the estimation procedure.

\section{Generalized Random Forests}\label{sec3}
Random Forests, introduced by \cite{Breiman2001}, are ensemble learning algorithms that construct a collection of decision trees using bootstrap sampling and random feature selection, aggregating tree-level predictions to reduce variance and improve robustness. While standard Random Forests target the conditional expectation
\begin{equation}
\mathbb{E}[Y \mid X = x], \label{eq:expectation}
\end{equation}
many applications require estimating quantities beyond conditional means, such as conditional quantiles, instrumental-variable parameters, or heterogeneous treatment effects. Generalized Random Forests (GRFs), proposed by \cite{Athey2019}, extend the Random Forest framework to estimate parameters defined via local estimating equations.

Let $\theta(x)$ denote the parameter of interest. Rather than minimizing prediction error directly, a GRF estimates $\theta(x)$ by solving the local moment condition
\begin{equation}
\theta(x) = \arg\min_{\theta} \mathbb{E} [\psi_\theta (O) \mid X = x] = 0, \label{eq:grf_moment}
\end{equation}
where $O$ denotes the observed data, $X$ the covariates, and $\psi_\theta (\cdot)$ a moment function. Equation (\ref{eq:grf_moment}) accommodates a range of statistical targets through the choice of $\psi$. Unlike conventional Random Forests, which partition observations to minimize squared prediction error, GRFs construct splits that maximize heterogeneity in the target parameter, enabling the identification of subpopulations with materially different parameter estimates.

\subsection{Treatment Effect Estimation}
For heterogeneous effect estimation, the target parameter is the Conditional Average Treatment Effect (CATE):
\begin{equation}
\tau(x) = \mathbb{E} [Y(1) - Y(0) \mid X = x] , \label{eq:cate}
\end{equation}
where $Y(1)$ and $Y(0)$ denote potential outcomes under treatment and control. Rather than assuming a constant effect $\tau(x) = \tau$, the GRF allows $\tau(x)$ to vary across the covariate space.

To reduce estimation bias, GRFs use honest estimation: each tree randomly partitions its bootstrap sample into a structure subset used to determine splits and a disjoint estimation subset used to compute within-leaf parameter estimates. Since no observation is used simultaneously for both structure and estimation, honest estimation improves the statistical validity and asymptotic properties of the resulting estimates \citep{Athey2019}.

The causal interpretation of $\tau(x)$ in Equation (\ref{eq:cate}) relies on standard assumptions for treatment effect estimation: unconfoundedness (treatment assignment is independent of potential outcomes given $X$), overlap (every covariate profile has a non-degenerate probability of receiving either treatment level), and the absence of interference between units (SUTVA). These conditions are satisfiable by design in a randomized experiment, but they are assumptions, not guarantees, in observational data, and they cannot be tested directly from $(X, W, Y)$ alone.

\section{Construction of a Pseudo-Treatment Variable}\label{sec4}
The BBBP dataset is observational and contains no experimentally assigned treatment. A binary indicator was therefore constructed from a biologically meaningful molecular property to allow the GRF machinery to be applied for exploratory purposes, while not claiming the resulting estimates are causal effects in the formal sense of Equation (\ref{eq:cate}).

The octanol-water partition coefficient ($\log P$) was selected as the basis for this pseudo-treatment. $\log P$ is defined as
\begin{equation}
\log P = \log_{10} \left( \frac{[C]_{\text{octanol}}}{[C]_{\text{water}}} \right), \label{eq:logp}
\end{equation}
where $[C]_{\text{octanol}}$ and $[C]_{\text{water}}$ are the equilibrium concentrations of a compound in octanol and water, respectively. $\log P$ quantifies molecular lipophilicity: higher values indicate greater affinity for lipid environments, while lower values indicate greater hydrophilicity. $\log P$ is established in the medicinal chemistry literature as a strong correlate of passive BBB permeability, since the barrier is composed primarily of lipid membranes \citep{Clark1999}; compounds with moderate lipophilicity generally permeate more readily, though excessive $\log P$ can reduce aqueous solubility and limit bioavailability.

Molecules were assigned to high- and low-lipophilicity groups using a median split:
\begin{equation}
W_i =
\begin{cases}
1, & \text{if } \text{LogP}_i > \text{Median}(\text{LogP}), \\
0, & \text{otherwise}.
\end{cases}
\label{eq:split_treatment}
\end{equation}

\section{Methodology}\label{sec5}
The selected feature space, comprising SMILES bigrams and molecular descriptors, formed the covariate matrix $X$. $\log P$ itself was excluded from $X$ to prevent information leakage into the treatment indicator, so that $W$ functioned exclusively as the (pseudo-)treatment and was not also available as a predictor.

A GRF was trained on $(X, W, Y)$ with honest splitting enabled, 2,000 trees, and a minimum leaf size of 10 observations. The forest produces a CATE estimate $\hat{\tau}(x)$ for each molecule together with an asymptotic confidence interval, and a normalized feature importance score for each covariate. The full estimation procedure, common to both the preliminary and orthogonalized analyses below, is summarized in Algorithm \ref{algo2}.


\begin{algorithm}
\caption{CATE Estimation with a Generalized Random Forest}\label{algo2}
\begin{algorithmic}[1]
\Require Training data $D = \{(x_i, y_i, \text{LogP}_i)\}_{i=1}^n$; feature matrix $X \in \mathbb{R}^{n \times p}$ (molecular descriptors and SMILES bigrams); binary outcome $Y \in \{0, 1\}$ (BBB permeability); continuous $\log P$ values; number of trees $B = 2000$; minimum node size $s = 10$ for honest splitting; significance level $\alpha = 0.05$
\Ensure CATE estimates $\hat{\tau}(x_i)$; confidence intervals $[\hat{\tau}_{\text{lower}}(x_i), \hat{\tau}_{\text{upper}}(x_i)]$; feature importance scores
\State Construct pseudo-treatment: compute $\text{LogP}_{\text{med}} = \text{median}(\text{LogP}_1, \dots, \text{LogP}_n)$ and set $W_i = \mathbb{I}[\text{LogP}_i > \text{LogP}_{\text{med}}]$; remove $\log P$ from $X$ to prevent leakage
\State Initialize forest: $B = 2000$ trees, minimum leaf size $s = 10$, honest splitting enabled, fixed random seed
\For{tree $b = 1, \dots, B$}
    \State Draw bootstrap sample $D_b$
    \State Randomly split $D_b$ into structure subset $D^{\text{split}}_b$ and estimation subset $D^{\text{est}}_b$ ($\approx 50/50$)
    \Repeat
        \State Select split $(j, c)$ on $D^{\text{split}}_b$ maximizing
        $\Delta(j, c) = \text{Var}(\hat{\tau}_{\text{split}}(\ell)) N_\ell + \text{Var}(\hat{\tau}_{\text{split}}(r)) N_r$
    \Until{node size $\leq s$}
    \State Using $D^{\text{est}}_b$, compute the within-leaf effect estimate:
    \[ \hat{\tau}_{\text{node}} = \frac{\sum_{i \in \text{node}}(W_i - \bar{W})(Y_i - \bar{Y})}{\sum_{i \in \text{node}}(W_i - \bar{W})^2} \]
\EndFor
\State Aggregate CATE estimates: $\hat{\tau}(x_i) = \frac{1}{B} \sum_{b=1}^B \hat{\tau}_b(x_i)$
\State Compute confidence intervals:
    \[ \widehat{\text{Var}}(\hat{\tau}(x)) = \frac{1}{B} \sum_{b=1}^B (\hat{\tau}_b(x) - \hat{\tau}(x))^2 \]
    \[ \hat{\tau}(x) \pm z_{1-\alpha/2} \sqrt{\widehat{\text{Var}}(\hat{\tau}(x))} \]
\State Flag significant heterogeneity: $\text{significant}_i = \mathbb{I}[0 \notin [\hat{\tau}_{\text{lower}}(x_i), \hat{\tau}_{\text{upper}}(x_i)]]$; report proportion of significant observations
\State Extract feature importances: $\text{Imp}_j = \frac{1}{B} \sum_{b=1}^B \sum_{\text{splits } s \text{ on } j} \Delta_s$, normalize to get $\widetilde{\text{Imp}}_j$
\State Rank covariates by $\widetilde{\text{Imp}}_j$
\end{algorithmic}
\end{algorithm}

The procedure was implemented using the \texttt{econml} Python library \citep{battocchi2019econml}. The preliminary analysis (Section \ref{subsec:naive}) used the \texttt{CausalForest} class directly; the primary analysis (Section \ref{subsec:dml}) used \texttt{CausalForestDML} to incorporate orthogonalized nuisance estimation, as detailed below.

Consistent with the framing in Section \ref{sec4}, $\hat{\tau}(x)$ is interpreted as the estimated difference in predicted BBB-permeability association between the above-median- and below-median-LogP groups, for a molecule with profile $x$, conditional on the model---not as the causal effect of manipulating $\log P$.

\section{Results and Discussion}\label{sec6}
\subsection{Treatment Group Balance}
The median-split construction produced a near-equal partition of the dataset (median $\log P$ cutoff: 0.076), with 815 molecules (49.97\%) in the high-LogP group ($W = 1$) and 816 molecules (50.03\%) in the low-LogP group ($W = 0$). This balance is a mechanical consequence of splitting on the sample median and confirms only that the two groups are similarly sized; it does not, by itself, provide evidence of overlap or unconfoundedness in the sense required for causal interpretation (Section \ref{sec3}).

\subsection{Preliminary Estimates: Naive Causal Forest}\label{subsec:naive}
As a first pass, and prior to any adjustment for confounding, a causal forest was fitted directly on $(X, W, Y)$ without orthogonalizing either the outcome or the treatment. This preliminary estimator produced a mean CATE of $\hat{\tau} = 0.121$ (SD $= 0.114$, range $[-0.101, 0.445]$) across the 1,631 molecules, with 46.9\% of estimates statistically significant at the 5\% level before multiple-testing correction. Feature importance under this estimator was dominated by broad physicochemical descriptors, led by TPSA (importance 0.364) and heteroatom count (0.202), together accounting for over 56\% of explained heterogeneity.

However, as noted in Section \ref{sec3}, the pseudo-treatment $W$ is constructed directly from $\log P$, a quantity that is itself strongly correlated with several covariates in $X$---including the very descriptors (TPSA, heteroatom count) that this preliminary estimator identifies as most important. This is precisely the confounding pattern that risks being misattributed to genuine treatment-effect heterogeneity, and motivates the orthogonalized re-estimation that follows. The preliminary result is therefore reported here as a baseline for comparison only, and should not be interpreted as evidence of genuine heterogeneity in isolation.

\subsection{Orthogonalized Estimation via Double Machine Learning}\label{subsec:dml}
To address the confounding identified above, the analysis was repeated using the \texttt{CausalForestDML} estimator, which embeds the causal forest within a Double Machine Learning (DML) framework \citep{Chernozhukov2018}, combined with the generalized random forest methodology of \cite{Athey2019} as described in Section \ref{sec3}.

Two nuisance models are estimated via cross-fitting: an outcome model $\hat{m}(X) = \mathbb{E}[Y \mid X]$, predicting BBB permeability from the molecular descriptors and SMILES-derived features alone, and a propensity model $\hat{e}(X) = \Pr(W = 1 \mid X)$, predicting group membership from the same covariates. Residualized outcome and treatment,
\[ \tilde{Y} = Y - \hat{m}(X), \quad \tilde{W} = W - \hat{e}(X), \]
are then passed to an honest causal forest in place of the original observations. Because both nuisance models are cross-fitted, the resulting estimator satisfies Neyman orthogonality, substantially reducing regularization bias while preserving asymptotic consistency and valid inference \citep{Chernozhukov2018, Athey2019}.

\subsubsection{Feature Importance}
Table \ref{tab3} presents the ten most influential variables identified by the orthogonalized causal forest.

\begin{table}[h]
\caption{Top ten variables explaining heterogeneous treatment effects estimated by the orthogonalized causal forest}\label{tab3}
\begin{tabular*}{\textwidth}{@{\extracolsep\fill}ll}
\toprule
Feature & Importance \\
\midrule
SMILES bigram \texttt{sc} & 0.1569 \\
TPSA & 0.0795 \\
SMILES bigram \texttt{=c} & 0.0460 \\
SMILES bigram \texttt{(o} & 0.0411 \\
Number of heteroatoms & 0.0395 \\
Minimum partial charge & 0.0328 \\
SMILES bigram \texttt{oc} & 0.0310 \\
Molar refractivity & 0.0250 \\
SMILES bigram \texttt{c[} & 0.0235 \\
Number of hydrogen bond donors & 0.0225 \\
\bottomrule
\end{tabular*}
\end{table}

Unlike the preliminary estimator, TPSA is no longer the dominant source of treatment-effect heterogeneity. The largest individual contribution is now provided by the SMILES bigram \texttt{sc}, and SMILES-derived features collectively account for an aggregate importance of 0.680. This shift is expected: the nuisance outcome model absorbs the variation in BBB permeability already explained by broad physicochemical descriptors before the causal forest is fit, leaving the forest to focus on residual structural information not captured by conventional descriptors. Feature importance here indicates variables that explain variation in the estimated treatment effects, not variables that directly determine BBB permeability.

SMILES bigrams warrant brief separate comment, as the dominant contributors in this orthogonalized result. Bigrams are generated by parsing each SMILES string into overlapping two-character subsequences (e.g., \texttt{CCO} yields \texttt{CC} and \texttt{CO}) and vectorizing them via TF-IDF, producing a sparse representation of local structural patterns such as ring closures or functional-group motifs. Their prominence here is consistent with several non-exclusive explanations: they may encode pharmacophoric substructure patterns not captured by explicit descriptors, supply residual structural information complementary to TPSA and heteroatom count, or capture context-dependent effects (e.g., a hydroxyl group behaving differently on an aromatic ring versus an aliphatic chain) that single-value descriptors cannot. Given the absence of statistically significant CATEs after multiple-testing correction (Section \ref{subsec:sig}), these potential explanations are speculative and would require external validation in independent datasets.

\subsubsection{Statistical Significance}\label{subsec:sig}
Statistical significance was evaluated using the asymptotic inference procedure provided by \texttt{econml}, with individual $p$-values adjusted via the Benjamini-Hochberg false discovery rate procedure \citep{Benjamini1995FDR}.

Before correction, 124 of the 1,631 estimated conditional treatment effects (7.60\%) were statistically significant at the 5\% level. This proportion is close to the 5\% expected by chance under the global null hypothesis of no heterogeneity, with the excess of 2.60 percentage points corresponding to approximately 42 additional nominally significant tests beyond what would be expected from random variation.

\begin{figure}[h]
    \centering
    \includegraphics[width=0.9\textwidth]{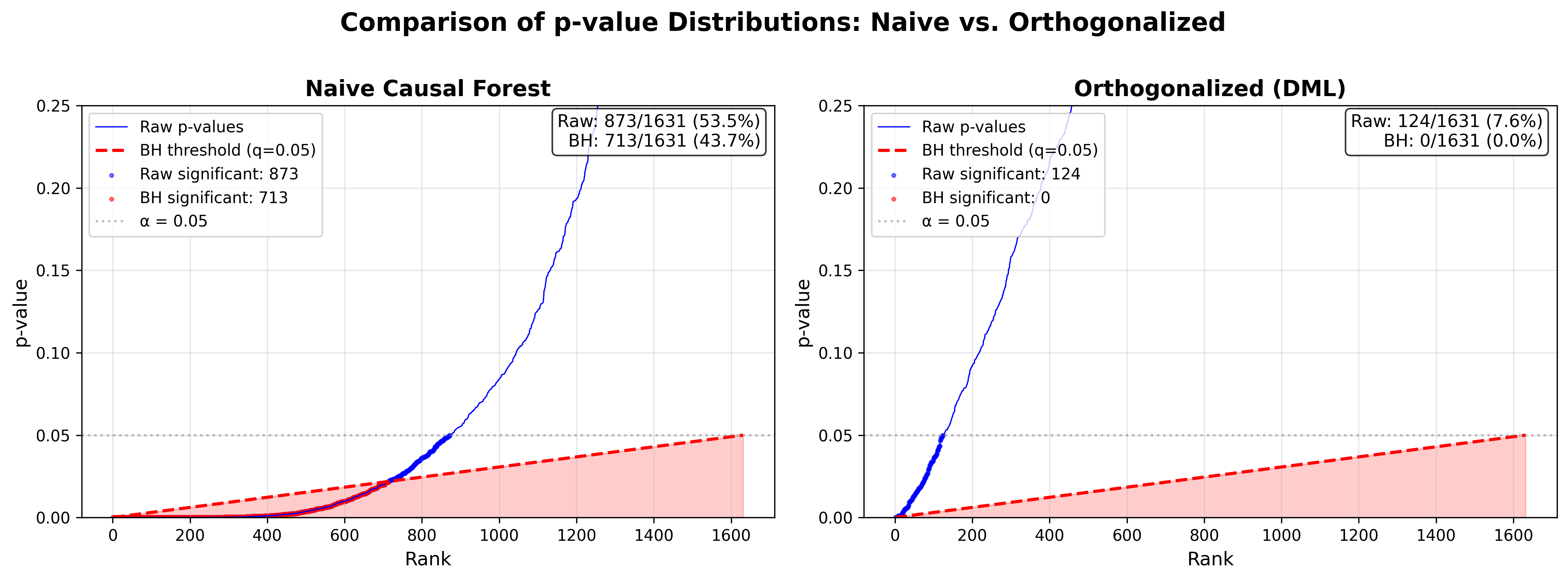}
    \caption{Comparison of p-value distributions and Benjamini-Hochberg correction for the naive and orthogonalized models. The red dashed line represents the BH threshold at $q = 0.05$. The naive model (left) shows 873 raw and 713 FDR-corrected significant findings; the orthogonalized model (right) shows 124 raw and 0 FDR-corrected significant findings, with the smallest adjusted $p$-value of 0.082.}\label{fig1}
\end{figure}

After controlling the false discovery rate ($q = 0.05$), none remained significant, with the smallest adjusted $p$-value equal to 0.082 (Figure \ref{fig1}). This finding is consistent with the interpretation that the observed nominal significances arose primarily from random variation rather than genuine treatment-effect heterogeneity.

The contrast with the naive estimator is stark (Table \ref{tab4}, Figure \ref{fig1}). The naive model suggested that 873 molecules (53.53\%) showed significant heterogeneity, and even after FDR correction, 713 molecules (43.72\%) remained significant, with many $p$-values effectively zero. This is a classic signature of severe confounding: the naive model attributes to treatment-effect heterogeneity the systematic associations between $\log P$ and molecular descriptors that influence BBB permeability.

\begin{figure}[h]
    \centering
    \includegraphics[width=0.9\textwidth]{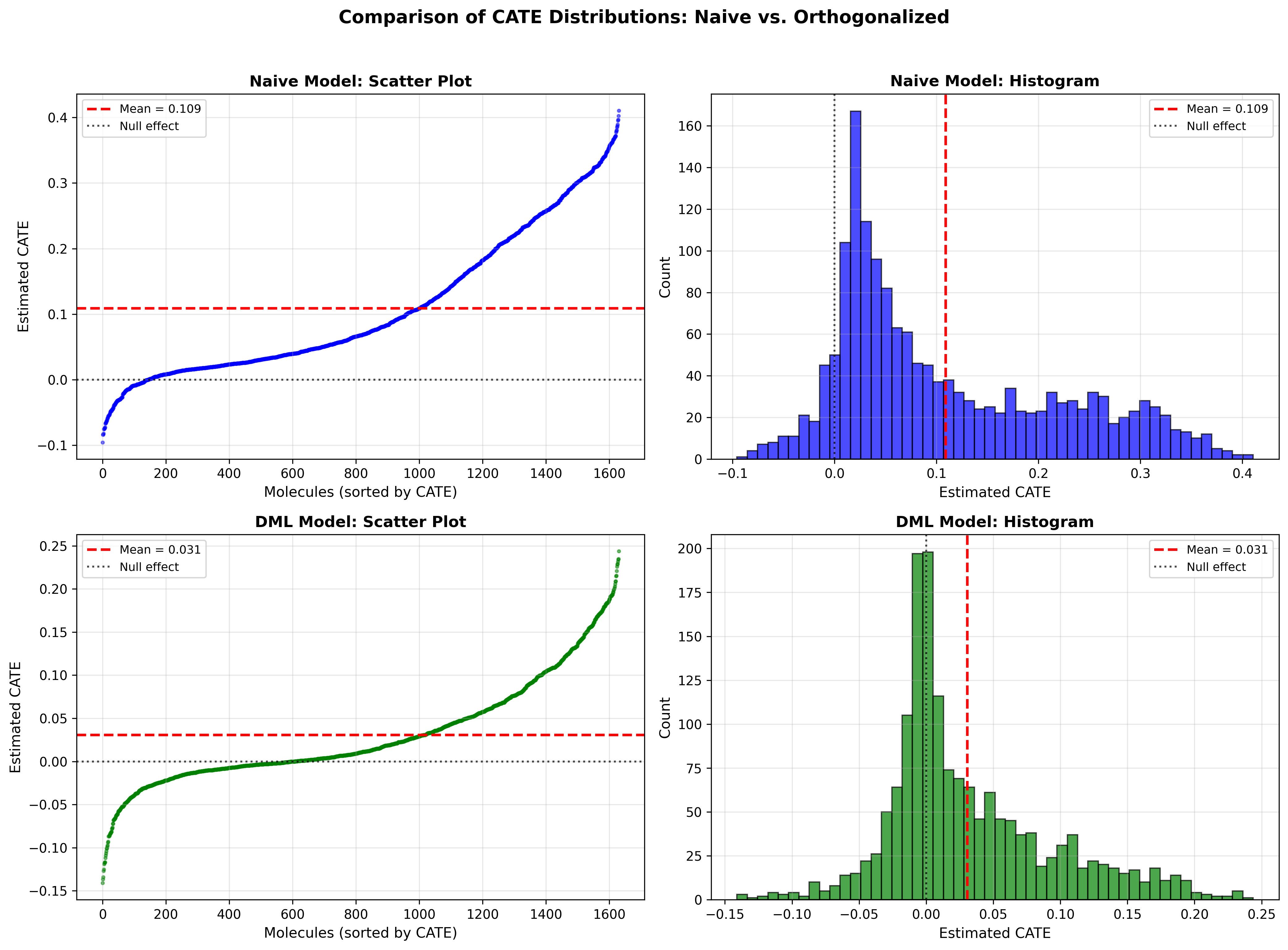}
    \caption{Comparison of Conditional Average Treatment Effect (CATE) distributions between the naive (top) and orthogonalized DML (bottom) estimators. The naive model shows substantial spread (mean = 0.109, std = 0.110, range = $[-0.096, 0.411]$) with many significant effects. The DML model shows tight clustering around zero (mean = 0.031), with no significant effects after FDR correction.}\label{fig2}
\end{figure}

The distribution of CATE estimates further illustrates this contrast (Figure \ref{fig2}). The naive model exhibits substantial spread, with CATEs distributed across a wide range of values. In contrast, the DML model shows tight clustering around zero, consistent with the absence of meaningful heterogeneity.

\begin{table}[h]
\caption{Comparison of naive and orthogonalized causal forest estimates}\label{tab4}
\begin{tabular*}{\textwidth}{@{\extracolsep\fill}lll}
\toprule
Metric & Naive Causal Forest & Orthogonalized (DML) \\
\midrule
Mean CATE & 0.109 & 0.031 \\
CATE standard deviation & 0.110 & 0.0616 \\
CATE range & $[-0.096, 0.411]$ & $[-0.1410, 0.2437]$ \\
Top feature & TPSA (0.305) & SMILES bigram \texttt{sc} (0.157) \\
Aggregate bigram importance & 0.382 & 0.680 \\
Raw significant ($p < 0.05$) & 873/1631 (53.53\%) & 124/1631 (7.60\%) \\
FDR-corrected ($q = 0.05$) & 713/1631 (43.72\%) & 0/1631 (0.00\%) \\
Smallest adjusted $p$-value & 0.000 & 0.082 \\
\bottomrule
\end{tabular*}
\end{table}

\subsection{Summary of Heterogeneity Findings}
The attenuation observed after orthogonalization indicates that much of the apparent heterogeneity detected by the preliminary estimator was attributable to systematic association between treatment assignment and the observed molecular descriptors, rather than genuine residual treatment-effect heterogeneity. The orthogonalized estimator explicitly removes outcome variation explained by measured descriptors and adjusts for the dependence between treatment assignment and covariates through cross-fitted nuisance models, yielding considerably more conservative and more credible estimates of heterogeneity.

The absence of statistically significant treatment effects after false discovery rate correction indicates that, once observed confounding is accounted for, there is no evidence that the molecular association between $\log P$ and BBB permeability varies systematically across compounds in this dataset. The apparent heterogeneity detected by the naive estimator was an artifact of confounding by molecular descriptors correlated with $\log P$. 

This null finding is itself informative: it demonstrates the importance of rigorous causal inference methods in chemoinformatics applications where treatment assignments are observational and confounded by molecular properties. Our results suggest that while $\log P$ is a well-known determinant of BBB permeability, its relationship with permeability is largely homogeneous across chemical space once other physicochemical properties are accounted for, at least within the scope of this dataset.

\section*{Availability of data and materials}
The datasets generated and/or analysed during the current study are available in the MoleculeNet benchmark repository. The code used for the generalized random forests methodology and feature ablation is available at \url{https://github.com/flash3105/Learning}.

\section*{CRediT authorship contribution statement}

\textbf{Tshemollo Rapolai:} Conceptualization, Methodology, Software, Formal analysis, Writing - original draft. \textbf{S. L. Makgai:} Supervision, Methodology, Validation, Writing - review \& editing. \textbf{M. Arashi:} Supervision, Methodology, Validation, Writing - review \& editing.

\section*{Declaration of competing interest}
The authors declare that they have no known competing financial interests or personal relationships that could have appeared to influence the work reported in this paper.

\section*{Declaration of generative AI and AI-assisted technologies in the manuscript preparation process}
During the preparation of this work, the author(s) used [ Claude, Gemini] in order to improve the language and readability of the manuscript. After using this tool/service, the author(s) reviewed and edited the content as needed and take(s) full responsibility for the content of the published article.

\nocite{*}
\bibliographystyle{elsarticle-num} 
\bibliography{refsfile}
\end{document}